\documentclass[11pt]{article}

\usepackage{amsmath, amssymb}
\usepackage{graphicx}
\usepackage{natbib}
\usepackage{geometry}
\usepackage{hyperref}
\usepackage{booktabs}
\usepackage{authblk}
\usepackage{caption}
\usepackage{float}

\title{Neural Networks Learn Distance Metrics}
\author{Alan Oursland}
\affil{\textit{alan.oursland@gmail.com}}
\date{February 2025}

\begin{document}

\maketitle

\begin{abstract}
    Neural networks may naturally favor distance-based representations, where smaller activations indicate closer proximity to learned prototypes. This contrasts with intensity-based approaches, which rely on activation magnitudes. To test this hypothesis, we conducted experiments with six MNIST architectural variants constrained to learn either distance or intensity representations. Our results reveal that the underlying representation affects model performance. We develop a novel geometric framework that explains these findings and introduce OffsetL2, a new architecture based on Mahalanobis distance equations, to further validate this framework. This work highlights the importance of considering distance-based learning in neural network design.
\end{abstract}
    
\section{Introduction}
\label{sec:introduction}

Neural networks have transformed machine learning through their ability to learn complex, hierarchical representations, traditionally interpreted through intensity-based representations where larger activation values signify stronger feature presence. This interpretation, originating from the McCulloch-Pitts neuron \cite{mcculloch1943logical} and Rosenblatt's perceptron \cite{rosenblatt1958perceptron}, underlies modern deep learning \cite{lecun2015deep}. However, our theoretical understanding of neural networks' internal mechanisms remains limited \cite{lipton2018mythos, montavon2018methods}, particularly regarding how they represent and process features.

In our prior work \cite{oursland2024interpreting}, we argued that networks may naturally learn distance-based representations, where smaller activations indicate proximity to learned prototypes. This reinterpretation challenges the traditional intensity-based paradigm and provides a statistical foundation rooted in principles like the Mahalanobis distance \cite{mahalanobis1936generalized, mclachlan2019mahalanobis}. Empirical evidence supports this theory. In our work \cite{oursland2024neural}, we demonstrated that distance-based metrics play a crucial role in how networks learn and utilize features, suggesting the need to rethink the fundamental nature of representations.

\textbf{Core Questions.} This paper investigates: (1) whether neural networks naturally prefer distance-based or intensity-based representations, (2) how architectural choices shape these representational biases, and (3) what geometric and statistical principles underlie these preferences.

\textbf{Contributions.} We examine neural network behavior through distance and intensity representations via: (1) A theoretical framework formalizing the distinction between these representations, (2) empirical analysis of six architectural variants, revealing mechanisms behind dead node creation and geometric performance limitations, and (3) introduction of OffsetL2, a novel architecture validating our framework through strong, stable performance.

The remainder of this paper reviews related work (Section \ref{sec:related_work}), establishes theoretical foundations (Section \ref{sec:background}), presents our experimental design (Section \ref{sec:exp_design}) and findings (Section \ref{sec:results}), and discusses implications (Section \ref{sec:discussion}).
\section{Related Work}
\label{sec:related_work}

Our work builds upon recent advances in understanding neural networks through statistical distance metrics. Our prior work \cite{oursland2024interpreting} demonstrated how linear layers with Absolute Value (Abs) activations approximate the Mahalanobis distance \cite{mahalanobis1936generalized}. We established a mathematical foundation for distance-based representations, showing that such layers naturally encode distance relationships rather than intensity-based activations. Empirical studies further reinforce this perspective. In \cite{oursland2024neural}, we demonstrated that networks with ReLU and Abs activations exhibit particular sensitivity to perturbations affecting distance relationships in the feature space, suggesting that distance metrics play a fundamental role in how networks process information.

Alternative approaches to incorporating distance metrics in neural networks include Radial Basis Function (RBF) networks \cite{broomhead1988radial, park1991universal}, which use distances from learned centers for classification, Siamese networks \cite{bromley1994signature, hadsell2006dimensionality}, which learn embeddings where distances represent similarity, and Learning Vector Quantization (LVQ) \cite{kohonen1995learning}, which explicitly models class prototypes and uses distance-based classification. Contrastive learning \cite{chen2020simple, he2020momentum} emphasizes the importance of learning representations where distances reflect semantic similarity, although those methods typically rely on carefully constructed training objectives rather than inherent architectural biases. While specialized architectures like RBF and LVQ demonstrate the effectiveness of explicitly encoding distances, they remain largely confined to specific applications rather than general-purpose deep learning.

Complementing these distance-centric views, geometric interpretations of neural computation offer valuable insights for understanding internal representations \cite{montavon2018methods, olah2017feature}. These approaches analyze hyperplanes and decision boundaries to explain how networks partition and represent data \cite{lipton2018mythos, erhan2009visualizing}, though they typically focus on networks trained under standard intensity-based assumptions.

This work bridges the gap between distance-based and geometric interpretations by investigating how architectural choices influence the emergence of distance-based representations.

\section{Background}
\label{sec:background}

This section introduces the theoretical framework for understanding how neural networks represent data through either distance-based or intensity-based methods, providing context for our experimental analysis.

\subsection{Features, Representations, and Distance Metrics}

Neural networks learn features through their internal representations. For the purposes of this paper, we define \textbf{features} as inherent properties of the data that can be quantified statistically, and \textbf{representations} are the compositions of these features learned by a node to generate its output. Traditional intensity-based representations encode features through activation magnitudes, while distance-based representations encode features through proximity to learned prototypes. In our discussion, we primarily consider prototypes as the features of interest, as they provide a structured way to quantify similarity and influence learned representations.

We argue that neural networks fundamentally learn distance features that quantify similarity between data points and learned prototypes. What appear to be intensity-based representations can be reinterpreted as disjunctive sets of distance features, corresponding to Disjunctive Normal Form (DNF) in Boolean algebra \cite{post1921introduction}. For example, an intensity representation for class $a$ implicitly learns $\lnot(b \lor c) = \lnot b \land \lnot c$, where $b$ and $c$ represent distances to other classes. This logical interpretation has connections to work exploring the relationship between neural networks and Boolean formulas \cite{anthony2003boolean}.

In contrast, distance-based representations directly encode proximity to prototypes as conjunctive sets (Conjunctive Normal Form). For class $a$, this simply requires a small distance to prototype $a$, represented logically as $a$. This more directly captures the underlying statistical relationships in the data.

\subsection{Distance vs. Intensity Representations}
\label{subsec:dist-intensity-rep}

Intensity-based interpretations, while lacking precise mathematical definition, are deeply embedded in the foundations of deep learning \cite{goodfellow2016deep}. These interpretations treat larger activations as indicating stronger feature presence, fundamentally shaping how we train and interpret networks. The use of one-hot encoded labels for classification directly encodes this assumption - the correct class should produce the highest activation while all others should be suppressed. This intensity-based paradigm appears throughout modern practice: cross-entropy loss encourages larger activations for correct classes, feature visualization \cite{erhan2009visualizing, olah2017feature} assumes maximum activations represent learned features, and even basic image processing interprets pixel intensities as direct measures of feature strength. However, this pervasive intensity-based interpretation remains an assumption imposed on neural networks rather than an inherent property.

Distance-based representations offer an alternative framework, interpreting smaller activations as indicating similarity to learned prototypes. This aligns with statistical metrics like the Mahalanobis distance \cite{mahalanobis1936generalized}, where the activation $f(x)$ is inversely proportional to the distance between input $x$ and prototypes $\mu$:

\begin{equation}
f(x) = |W(x - \mu)|_p,
\end{equation}

where $W$ is a learned scaling matrix and $p$ denotes the norm. Common activation functions naturally support this view: the absolute value function directly represents distance, while ReLU implicitly encodes distance through the relationship $Abs(x) = ReLU(x) + ReLU(-x)$. This framework emphasizes geometric relationships in the latent space rather than activation magnitudes. This perspective aligns with research on visualizing and understanding the loss landscape of neural networks \cite{li2018visualizing, goodfellow2015qualitatively}.

\subsection{Theoretical Foundations: Mahalanobis Distance}

The Mahalanobis distance provides a statistical foundation for distance-based representations, capturing feature correlations and scales through the covariance matrix $\Sigma$:
\begin{equation}
    D_M(x) = \sqrt{(x - \mu)^T \Sigma^{-1} (x - \mu)},
\end{equation}

Through eigendecomposition $\Sigma = V\Lambda V^T$, this distance can be expressed as:
\begin{equation}
    D_M(x) = \| \Lambda^{-1/2} V^T (x - \mu) \|_2,
\end{equation}
revealing how neural networks naturally approximate this metric: linear layer weights learn the eigenvector transformation $\Lambda^{-1/2} V^T$, the bias learns $-\Lambda^{-1/2} V^T \mu$, and activation functions like Absolute Value model the normalized distance computation \cite{oursland2024interpreting}.

\subsection{Geometric Interpretations of Representations}

Neural networks project input data into a latent space structured by hyperplanes; under a distance-based interpretation, these hyperplanes act as axes of a manifold, with distances to prototypes along these axes defining the representation \cite{montavon2018methods}.

For a hyperplane defined by $y = Wx + b$, where $x$ is $N$-dimensional and $y$ is scalar, its decision boundary occurs where the activation function equals zero (e.g., $0 = Wx + b$). This boundary is uniquely described by $N$ linearly independent points: $N-1$ points on the decision boundary plus one point at ${x=0, y=b}$. These boundary points serve as feature prototypes, encoding relationships between inputs and their latent representations. These prototypes don't necessarily reflect human-perceived similarity, but rather capture relationships defined by the network's learned distance metric.

When $b=0$, hyperplanes must pass through the origin, making it a fixed prototype. This constraint has minimal impact in high dimensions since each hyperplane still intersects $N-2$ independent points on its decision boundary, plus points at the origin and ${x!=0, y!=0}$. While these prototypes are theoretically significant, recovering them becomes computationally intractable as dimensionality increases \cite{donoho2000high}.

\subsection{Activation Functions and Representational Preferences}

Activation functions shape how networks encode representations:

\begin{itemize}
    \item \textbf{ReLU}: $f(x) = \max(0, x)$ is traditionally associated with intensity-based interpretations but can be viewed through a distance lens \cite{nair2010relu, glorot2011deep}. The zero region corresponds to one side of a decision boundary, with positive activations encoding prototype proximity. However, ReLU's tendency to produce "dead neurons" when inputs remain negative can hinder learning. This issue has motivated research into alternative activation functions \cite{he2015delving, ramachandran2017searching, misra2019mish}.
    
    \item \textbf{Absolute Value (Abs)}: $f(x) = |x|$ directly represents distance-based relationships by preserving magnitude information regardless of sign. This symmetry ensures neurons remain active and maintain complete information about distance from decision boundaries.
    
    \item \textbf{Neg Layers}: $f(x) = -x$ transforms positive distance representations into negative intensity representations by inverting activation order.
\end{itemize}

This theoretical foundation frames our experimental analysis, where we systematically investigate how architectural choices, such as activation functions and bias terms, affect representational biases. By probing these factors, we aim to elucidate the fundamental principles driving neural network learning and provide a unified framework for understanding distance-based and intensity-based representations.

\section{Experimental Design}
\label{sec:exp_design}

We explore whether networks exhibit a preference for distance-based or intensity-based representations. We employ simple two-layer networks and systematically architectural components to force either distance or intensity representations in the final linear layer. We aim to reveal potential training biases towards specific representation types.

\subsection{Objectives}
\begin{enumerate}
    \item When neural networks are constrained to produce either distance-based or intensity-based outputs, how does this affect their performance?
    
    \item How do different activation functions influence the network's ability to learn under these different representational constraints?
    \item By analyzing the performance and behavior of networks under these constraints, what can we infer about the nature of the representations learned in the output layer and the geometry of the feature space?
\end{enumerate}

\subsection{Model Design}

We constrained six neural network designs to force specific internal representations. These architectures are intentionally kept simple to isolate the specific behaviors under study. We utilize two-layer networks. All of the hidden layers have 128 nodes. ReLU is studied as a standard activation function in deep learning. Abs is studied for its theoretical connection to the Mahalanobis distance.

The core representational constraint is CrossEntropyLoss, which enforces intensity-based outputs. The combination of activations and a negation operator controls how the output layer must represent features internally. The activation function forces positive values (more precisely, non-negative). The negation reverses the order and signs of the activations. Models with the negation learn a positive distance representation which is converted by the negation layer into a negative intensity representation. 

The six primary architectures exclude a bias term in the final linear layer. This choice prevents the network from trivially learning the opposite representation (because $-(Wx)=(-W)x$) and then simply shifting it back to the positive side by using the bias. 

To establish a baseline for comparison, we include two control architectures \texttt{ReLU} and \texttt{Abs}. The four experimental architectures are \texttt{Abs2}, \texttt{Abs2-Neg}, \texttt{ReLU2}, and \texttt{ReLU2-Neg}. 

\begin{table}[H]
    \centering
    \begin{tabular}{ll}
        \toprule
        \textbf{Model} & \textbf{Architecture}\\
        \midrule
        Abs & x → Linear → Abs → Linear → y \\
        ReLU & x → Linear → ReLU → Linear → y \\
        \midrule
        Abs2 & x → Linear → Abs → Linear → Abs → y \\
        Abs2-Neg & x → Linear → Abs → Linear → Abs → Neg → y \\
        ReLU2 & x → Linear → ReLU → Linear → ReLU → y \\
        ReLU2-Neg & x → Linear → ReLU → Linear → ReLU → Neg → y \\
        \bottomrule
    \end{tabular}
    \caption{Experiment Model Architectures}
\end{table}

\subsection{Experimental Setup}
We use the MNIST dataset \cite{lecun1998gradient} for its well-understood features and relatively low dimensionality (28x28 pixels), making it suitable for analyzing representational preferences in a controlled setting. As is standard practice, images are normalized to zero mean and unit variance across the dataset.

To minimize confounding factors, we choose a simple training protocol with minimal hyperparameters. We use SGD optimization with learning rate 0.001 and train for 5000 epochs with full-batch updates. Each experiment is repeated 20 times. The loss function is CrossEntropyLoss (which includes LogSoftmax) applied to the final layer's logits.

We evaluate each architecture's performance using three metrics: test accuracy on MNIST, stability (variance across 20 training runs), and statistical significance via paired t-tests between architectures. These metrics enable us to compare both absolute performance and the consistency of learning across different random initializations.

Source code and additional resources can be found in our GitHub repository\footnote{\url{https://github.com/alanoursland/neural_networks_learn_distance_metrics}}.

The performance of each architecture, under the described experimental conditions, is analyzed in the following section. 

\section{Experimental Results}
\label{sec:results}

We conducted extensive experiments comparing baseline architectures and architectural variants. All experiments were run for 5,000 epochs with 20 independent trials to ensure statistical robustness.

\begin{table}[H]
    \centering
    \begin{tabular}{lcc}
        \toprule
        \textbf{Model} & \textbf{Test Accuracy (\%)} & \textbf{Standard Deviation (\%)}\\
        \midrule
        Abs & 95.87 & 0.22 \\
        ReLU & 96.62 & 0.17 \\
        \midrule
        Abs2 & 95.95 & 0.17 \\
        Abs2-Neg & 92.25 & 2.07 \\
        ReLU2 & 56.31 & 19.31 \\
        ReLU2-Neg & 96.46 & 0.17 \\
        \bottomrule
    \end{tabular}
    \caption{Performance metrics across all model variants after 5,000 epochs of training. Results show mean test accuracy, standard deviation, 95\% confidence intervals, and number of independent trials.}
    \label{tab:model_performance}
\end{table}

\subsection{Baseline Performance}
The baseline \texttt{ReLU} and \texttt{Abs} architectures showed strong performance on MNIST, with no statistically significant difference between them (t(38) = 1.14, p = 0.26, Cohen's d = 0.37). This comparable performance suggests that the choice of activation function alone does not significantly impact model effectiveness under standard conditions. These results provide a robust foundation for evaluating our architectural modifications.

\subsection{Intensity Learning Models}

The models constrained to learn intensity representations through the addition of a second activation function, \texttt{ReLU2} and \texttt{Abs2}, exhibited markedly different behaviors. 

\texttt{ReLU2}'s performance degraded catastrophically, showing a substantial drop from the baseline ReLU model (t(38) = -17.33, $p < 0.001$, Cohen's d = 5.56). This dramatic failure aligns with our hypothesis that neural networks may exhibit a bias towards learning distance-based representations. 

In contrast, \texttt{Abs2} maintained performance statistically indistinguishable from the baseline Abs model (t(38) = 1.4967, p = 0.1427, Cohen's d = 0.47). This finding complicates our initial hypothesis, suggesting that the relationship between activation functions and representational biases may be more nuanced than initially theorized.

\subsection{Distance Learning Models}

The models designed to learn distance representations through the Negation layer, \texttt{ReLU2-Neg} and \texttt{Abs2-Neg}, showed contrasting behaviors.

\texttt{ReLU2-Neg} exhibited a remarkable recovery from \texttt{ReLU2}'s catastrophic failure (t(38) = -17.33, p < 0.001, Cohen's d = 5.48), achieving performance statistically comparable to the baseline \texttt{ReLU} (t(38) = -12.78, p < 0.001, Cohen's d = 4.04). This recovery supports our hypothesis that neural networks may be biased towards learning distance-based representations, with the Neg transformation enabling \texttt{ReLU2-Neg} to leverage this bias effectively.

Surprisingly, \texttt{Abs2-Neg} showed significant performance degradation compared to both the baseline \texttt{Abs} (t(38) = -8.81, p < 0.001, Cohen's d = 2.79) and its intensity counterpart, \texttt{Abs2} (t(38) = 8.97, p < 0.001, Cohen's d = 2.84). The markedly higher variability in \texttt{Abs2-Neg}'s performance (SD = 2.56\% vs. 0.17\% for \texttt{Abs2}) further suggests that enforcing distance-based learning through negation may fundamentally interfere with the Abs activation function's learning dynamics.

\subsection{Impact of Bias Exclusion}

We excluded the bias term from the second linear layer to enforce learning through the origin, effectively reducing the dimensionality of the solution space by one. For completeness, we conducted parallel experiments with the bias term included (Table~\ref{tab:biased_performance}).

\begin{table}[H]
    \centering
    \begin{tabular}{lcc}
        \toprule
        \textbf{Model} & \textbf{Test Accuracy (\%)} & \textbf{Standard Deviation (\%)} \\
        \midrule
        Abs\_Bias & 95.23 & 0.16 \\
        ReLU\_Bias & 95.69 & 0.17 \\
        \midrule
        Abs2\_Bias & 95.38 & 0.17 \\
        Abs2\_Neg\_Bias & 90.53 & 2.36 \\
        ReLU2\_Bias & 39.94 & 18.84 \\
        ReLU2\_Neg & 94.92 & 0.18 \\
        \bottomrule
    \end{tabular}
    \caption{Performance metrics of models with bias terms included.}
    \label{tab:biased_performance}
\end{table}

The inclusion of bias terms had minimal impact on the overall patterns observed in our main experiments. \texttt{ReLU2\_Bias} maintained poor performance and high variance, while \texttt{Abs2\_Bias} and \texttt{Abs2\_Neg\_Bias} preserved their relative performance characteristics. These results suggest that the representational biases we observed are robust to the inclusion of bias terms and stem from more fundamental aspects of the architectures.

\subsection{Summary of Findings}

The experiments revealed that seemingly minor architectural changes can significantly impact model performance, yielding both expected and surprising results. The catastrophic failure of \texttt{ReLU2} under intensity constraints aligned with our predictions about distance-based representational bias. However, \texttt{Abs2}'s resilience to these same constraints complicated this narrative. The distance-constrained models further nuanced our understanding: \texttt{ReLU2-Neg}'s recovery to baseline performance supported our distance-bias hypothesis, while \texttt{Abs2-Neg}'s significant underperformance revealed unexpected limitations.

These contrasting behaviors suggest that neural networks' representational capabilities are more nuanced than our initial hypothesis predicted. While networks can adopt both distance- and intensity-based approaches, their success appears highly dependent on the specific architectural configuration, particularly the choice of activation function. This interplay between architecture and representation forms the focus of our subsequent geometric analysis in the Discussion section.

\section{Discussion}
\label{sec:discussion}

Our experiments revealed unexpected behaviors in how neural networks learn representations. While we hypothesized a preference for distance-based representations, the results paint a more complex picture: \texttt{ReLU2} failed catastrophically when constrained to learn intensity representations, yet \texttt{Abs2} showed surprising resilience to these same constraints. Meanwhile, \texttt{Abs2-Neg} underperformed despite being designed for distance-based learning. These counterintuitive findings suggest that the relationship between network architecture and representational capacity is more nuanced than initially theorized. This aligns with research highlighting the complex interplay between architecture, optimization, and generalization in deep learning \cite{bengio2013representation, jacot2018neural, lee2019wide}.

\subsection{Feature Distributions in Latent Spaces}

To help visualize our geometric analysis, we analyze how data points are distributed relative to the hyperplanes defined by the first linear layer, using preactivation values to measure distances from decision boundaries. Since precise feature identification is challenging in deep networks, we use MNIST class labels as proxies to understand how different classes cluster in the latent space.

\begin{figure}[H]
    \centering
    \includegraphics[width=0.8\textwidth]{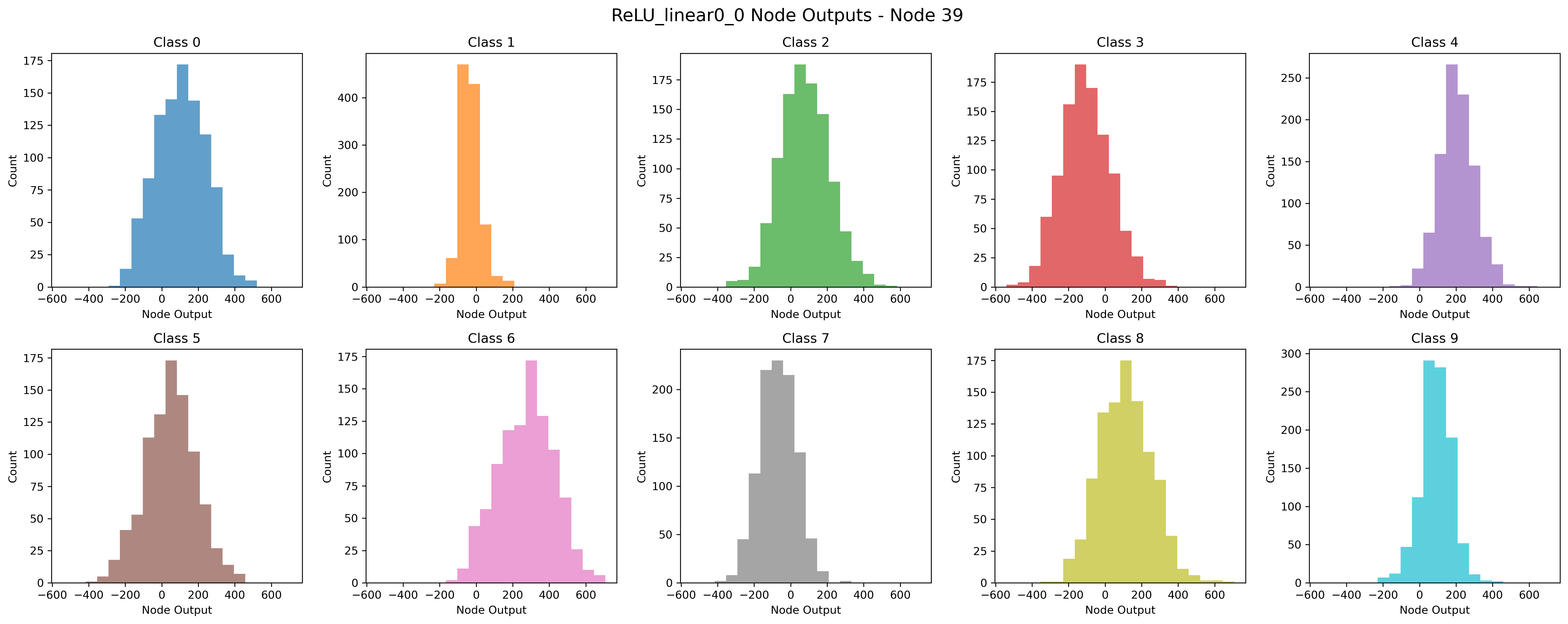}
    \caption{Class distributions in the latent space show overlapping clusters with varying statistical properties. Each class exhibits distinct characteristics (mean, variance, skewness), with overlap patterns varying across different linear projections - typical behavior for non-linearly separable data.}
    \label{fig:distance_distribution_label}
\end{figure}

The first linear layer's outputs form a 128-dimensional latent space, where the second layer defines a hyperplane $y=Wx+b$ ($b=0$). While traditional analysis views hyperplanes as boundaries that separate clusters, our distance-based interpretation focuses on how hyperplanes intersect clusters to define prototypes \cite{snell2017prototypical, weinberger2009distance}. The hyperplane is uniquely defined by 128 points: 127 points on the decision boundary (representing learned prototypes within clusters), the origin (due to $b=0$), and one point outside the latent space. These intersections with clusters determine the hyperplane's orientation and thus how the network measures distances to prototypes for classification.

In this latent space, we define two important points for each class $c$: an optimal center $z_c$ that minimizes intra-class distances while maximizing inter-class distances, and its counterpart $z_{\neg c}$ that does the opposite. These points, constructed from the first layer's node outputs, represent ideal prototypes within their respective clusters. This formulation aligns with prior work on structuring feature spaces through learned metric embeddings \cite{weinberger2009distance, xing2002distance}.

When learning a distance representation, the second linear layer's hyperplane attempts to intersect points approximating $z_c$ on its decision boundary \cite{oursland2024interpreting}. This alignment ensures the target class has minimal activations by positioning the boundary near its ideal center, effectively identifying classes based on their proximity to these centers. 

When learning an intensity representation, the second linear layer's hyperplane attempts to intersect points approximating $z_{\neg c}$ on its decision boundary. This alignment ensures the target class has maximal activations by positioning the boundary near the centers of non-target classes, effectively separating classes based on their distance from these "anti-centers."

\subsection{Analysis of ReLU-based Architectures}

\texttt{ReLU2} failed catastrophically ($47.20\% \pm 12.00\%$) due to widespread node death in its output layer: $33.00\%$ of nodes were permanently inactive and $53.50\%$ activated for less than 5\% of inputs. This failure stems from attempting to learn a disjunctive distance representation while constrained by intensity-based learning. Since non-target classes comprise 90\% of the data for any classification decision, the network drives most pre-activations negative to minimize non-target activations. This issue of inactive or dying ReLUs has been widely studied in the context of rectifier activations \cite{he2015delving}.

The dead node collapse emerges from a compounding effect: when classes overlap in the latent space, minimizing activations for non-target classes ($\neg c$) inevitably affects the target class ($c$) as well. Since each class comprises only 10\% of the data, the optimization overwhelmingly prioritizes minimizing non-target activations (90\% of cases) over preserving activations for the target class (10\% of cases). As a result, pre-activations for both target and non-target classes are driven negative, which the second ReLU then zeros out, leading to widespread dead nodes.

In contrast, \texttt{ReLU2-Neg} achieves near-baseline performance ($94.93\% \pm 0.15\%$) by building a conjunctive distance representation. It positions hyperplanes so that class $c$ points have negative pre-activations (centered around $z_c$), which ReLU converts to zero. Crucially, this also applies to all classes with even smaller pre-activations (i.e., those positioned to the left of $z_c$ in the projected space, as shown in Figure~\ref{fig:distance_distribution_label}). Since ReLU zeros out everything below the target class, the network must rely on the decorrelation of these classes across different hyperplane projections to prevent them from overlapping with the target class. The diverse projections in the latent space \cite{bengio2013representation, cogswell2016decorrelating} ensure this decorrelation, allowing the target class to maintain minimal activation while maximizing $\neg c$.

\subsection{Analysis of Abs-based Architectures}

Unlike ReLU, Abs networks cannot produce dead nodes. Instead of zeroing out negative values, Abs folds them to the positive side, ensuring all nodes remain active. Under our distance metric theory—where zero activation signifies maximum feature membership—this architectural difference leads to distinct feature representations. 

In ReLU networks, maximum feature membership extends to all input regions producing negative pre-activations, creating broader feature sets. In contrast, Abs networks achieve maximum membership only at exact decision boundary points, resulting in more focused feature sets. Since the minimum activation can correspond to either $z_c$ or $z_{\neg c}$, Abs networks provide a more direct encoding of distances to learned prototypes or anti-prototypes.

While \texttt{Abs2} performed well ($95.35\% \pm 0.17\%$), its distance-learning counterpart, \texttt{Abs2-Neg}, suffered a notable accuracy drop ($90.08\% \pm 2.56\%$) with significantly higher variance. What explains this unexpected performance gap?

We theorize that \texttt{Abs2-Neg} underperformance may be related to the clustered nature of MNIST \cite{deng2012mnist, perez2017effectiveness}. This dataset's distinct clusters might lead to the existence of a single, highly optimal prototype point, $z_c$, for each class. An output hyperplane in \texttt{Abs2-Neg} must pass through that optimal prototype $z_c$ and 127 additional linearly independent points. If a single, dominant $z_c$ exists, the remaining points must be suboptimal, potentially lying closer to non-target class distributions. This constraint could lead to misclassifications and higher variance. The development of \texttt{OffsetL2}, which explicitly models a single prototype per class, was motivated by this hypothesis.

In contrast, \texttt{Abs2} constructs its hyperplane by selecting $z_{\neg c}$, the centers of non-target classes, for each latent dimension. Since each dimension can be aligned with any of the nine non-target classes, \texttt{Abs2} has a vast combinatorial space of possible hyperplane configurations ($9^{128}$ choices). This flexibility allows it to compensate for suboptimal points in some dimensions by making better choices in others. As a result, \texttt{Abs2} can achieve robust class separation, explaining its higher accuracy and lower variance compared to \texttt{Abs2-Neg}.

\subsection{Validation Through Additional Experiments}

Our theory about the \texttt{Abs2-Neg} performance drop suggests that a layer designed to explicitly represent the distance to a single optimal point might correct the performance difference. To address the limitations of \texttt{Abs2-Neg}, where the need for multiple non-optimal intersection points hampered performance, we propose a layer called OffsetL2 that computes the weighted L2 distance from a single learned reference point $\mu$:

\[
y_i = || \alpha_i \odot (x - \mu_i) ||_2
\]

OffsetL2 directly implements our geometric intuition by explicitly learning a single optimal reference point, $\mu_i$, for each class, corresponding to the hypothesized ideal prototype $z_c$. The learnable weight vector $\alpha_i$ modulates the importance of each dimension in the distance calculation, providing greater flexibility. This approach contrasts with \texttt{Abs2-Neg}, which implicitly discovers prototypes through hyperplane positioning. This is conceptually similar to the learned prototypes in Radial Basis Function (RBF) networks, where classification is determined by distance to a set of learned reference points \cite{moody1989fast}, and to the Mahalanobis distance formulation discussed in the Background.

\begin{table}[H]
    \centering
    \begin{tabular}{lc}
        \toprule
        \textbf{Method} & \textbf{Equation} \\
        \midrule
        OffsetL2 + LogSoftmax & $ y_i = \exp(-||\alpha_i \odot (x - \mu_i) ||_2) $ \\  
        Traditional RBF & $ y_i = \exp(-0.5 (\text{precision}_i (x - \mu_i)^2)) $ \\  
        Mahalanobis + LogSoftmax & $ y_i = \exp(-||\text{precision}_i v_i (x - \mu_i)||_2) $ \\  
        \bottomrule
    \end{tabular}
    \caption{Comparison of OffsetL2 with related distance-based methods.}
    \label{tab:comparison_offsetl2}
\end{table}

When preceded by a linear layer, OffsetL2 becomes functionally equivalent to the PCA-based Mahalanobis distance, where the linear layer learns principal components ($V$) and OffsetL2 learns the scaling ($\Lambda^{-1/2}$) and mean ($\mu$). This strong connection between OffsetL2, RBF networks, and Mahalanobis distance further reinforces its theoretical grounding.

To evaluate OffsetL2, we introduced four new models: \texttt{ReLU-L2}, \texttt{ReLU-L2-Neg}, \texttt{Abs-L2}, and \texttt{Abs-L2-Neg}. Training was extended to 50,000 epochs after observing that models had not fully converged at 5,000 epochs.

\begin{table}[H]
    \centering
    \begin{tabular}{lcc}
    \toprule
    \textbf{Model} & \textbf{Accuracy (\%)} & \textbf{Std Dev (\%)} \\
    \midrule
    ReLU\_Bias & 96.62 & 0.17 \\
    ReLU2\_Bias & 56.31 & 19.31 \\
    ReLU2\_Neg & 96.46 & 0.17 \\
    Abs\_Bias & 95.87 & 0.22 \\
    Abs2\_Bias & 95.95 & 0.17 \\
    Abs2\_Neg\_Bias & 92.25 & 2.07 \\
    \midrule
    ReLU-L2 & 97.33 & 0.13 \\
    ReLU-L2-Neg & 97.36 & 0.14 \\
    Abs-L2 & 97.61 & 0.07 \\
    Abs-L2-Neg & 97.56 & 0.09 \\
    \bottomrule
    \end{tabular}
    \caption{Performance metrics across all models with extended training (50,000 epochs), averaged over 20 runs.}
    \label{tab:extended_training}
\end{table}

The results demonstrate several key findings:

1. The performance gap between normal and negated variants disappeared, supporting our theory about explicit prototype learning.
2. \texttt{ReLU-L2} avoided the catastrophic failure of \texttt{ReLU2\_Bias}.
3. OffsetL2 architectures significantly outperformed baselines, with \texttt{Abs-L2} models achieving $\sim 97.6\%$ accuracy.
4. All OffsetL2 models exhibited remarkably low variance ($\leq 0.14\%$).

These findings validate our theoretical framework: by explicitly modeling geometric constraints through direct distance calculations, OffsetL2 not only improves accuracy but also stabilizes training. The convergence in performance between normal and negated variants provides strong empirical validation of our core hypothesis—explicitly modeling distances to learned prototypes leads to more robust and accurate learning.

Our results, combined with geometric analysis, suggest a fundamental shift in how neural network representations should be conceptualized. Moving beyond the traditional intensity-based paradigm, these findings highlight the power of statistical distance metrics and geometric constraints, paving the way for new architectural advances in deep learning.

\section{Conclusion}
\label{sec:conclusion}

Our analysis reveals fundamental insights into the geometric principles underlying neural network representations through the lens of statistical distances. We demonstrate that while networks can adopt either representation, ReLU-based architectures exhibit a natural bias towards distance-based learning. The catastrophic failure of \texttt{ReLU2} under intensity constraints illustrates how architectural choices can create untenable optimization landscapes, particularly when inputs must cluster near decision boundaries. The performance gap between \texttt{Abs2} and \texttt{Abs2-Neg} further illuminates this geometric perspective: while \texttt{Abs2} leverages combinatorial flexibility to select optimal separation points in high-dimensional space, \texttt{Abs2-Neg}'s restricted prototype selection leads to degraded performance. This distinction underscores the crucial role of architectural flexibility in learning effective distance-based representations.

These findings suggest that network behavior is driven by geometric interactions in the feature space rather than intrinsic properties of activation functions. The success of our OffsetL2 architecture, which directly models statistically-motivated geometric relationships through Mahalanobis distance calculations, validates this framework by achieving superior performance (97.61\% accuracy) with remarkable stability ($\pm$ 0.07\% standard deviation). By learning single prototypes representing either optimal $z_c$ or $z_{\neg c}$ for each class, OffsetL2 avoids the pitfalls of implicit prototype discovery through constrained hyperplanes, demonstrating the advantages of explicitly modeling distance-based relationships.

This research opens new avenues for neural network design by demonstrating the importance of explicitly modeling geometric relationships in feature space. Future work should explore how these principles extend to deeper architectures and diverse tasks, potentially leading to networks that are not only more powerful but also more interpretable and aligned with the underlying statistical structure of the data. By viewing neural computation through the lens of statistical distances rather than activation intensities, we can develop more principled approaches to architecture design that bridge the gap between theoretical understanding and practical application.

\bibliographystyle{plainnat}
\bibliography{references}

\end{document}